\documentclass[sigconf]{acmart}
\usepackage{placeins} 

\usepackage{setspace}
\usepackage{float}

\setcopyright{acmcopyright}
\copyrightyear{}
\acmYear{}
\acmDOI{}
\acmISBN{}

\DeclareMathOperator*{\argmax}{arg\,max}

\begin{document}

\title{Calibrated Recommendations with Contextual Bandits}

\author{Diego Feijer}
\affiliation{%
  \institution{Spotify}
}

\author{Himan Abdollahpouri}
\affiliation{%
  \institution{Spotify}
}

\author{Sanket Gupta}
\affiliation{%
  \institution{Spotify}
}
\author{Alexander Clare}
\affiliation{%
  \institution{Spotify}
}

\author{Yuxiao Wen}
\affiliation{%
  \institution{NYU}
}

\author{Todd Wasson}
\affiliation{%
  \institution{Spotify}
}
\author{Maria Dimakopoulou}
\affiliation{%
  \institution{Uber}
}

\author{Zahra Nazari}
\affiliation{%
  \institution{Spotify}
}
\author{Kyle Kretschman}
\affiliation{%
  \institution{Spotify}
}
\author{Mounia Lalmas}
\affiliation{%
  \institution{Spotify}
}

\renewcommand{\shortauthors}{}

\begin{abstract}

Spotify’s Home page features a variety of content types, including music, podcasts, and audiobooks. However, historical data is heavily skewed toward music, making it challenging to deliver a balanced and personalized content mix. Moreover, users' preference towards different content types may vary depending on the time of day, the day of week, or even the device they use. We propose a calibration method that leverages contextual bandits to dynamically learn each user’s optimal content type distribution based on their context and preferences. Unlike traditional calibration methods that rely on historical averages, our approach boosts engagement by adapting to how users interests in different content types varies across contexts. Both offline and online results demonstrate improved precision and user engagement with the Spotify Home page, in particular with under-represented content types such as podcasts.

\end{abstract}



\maketitle

\section{Introduction}

Many streaming platforms offer a diverse range of content types to cater to a broad spectrum of user interests. For instance, Spotify offers music, podcasts, and audiobooks. As platforms continue to expand their content offerings, it becomes increasingly important for recommender systems and ranking algorithms to effectively balance these different content types in accordance with each user's individual preferences to preserve long-term user satisfaction \cite{mcnee2006being}, prevent the narrowing-down of a user’s interests that can hinder discovery and exploration \cite{castells2021novelty}, and mitigate popularity biases \cite{klimashevskaia2024survey} that favor more popular and established content types. 

A widely used approach in industry for diversifying recommendations based on user preferences is \textit{calibration}, as introduced by \cite{steck2018calibrated}. This  post-processing {\it re-ranking} method operates on a large initial set of candidate items, combining the relevance scores of items with a diversity score of the recommended list measured via Kullback–Leibler (KL)-divergence, and constructs the list iteratively to build recommendation sets that are both relevant and calibrated to the user's taste across different content types. For example, \cite{steck2018calibrated} applied calibration to diversify movie recommendations by genres, aligning the distribution of genres in the recommendation set with each user's individual genre preferences. Similarly, \cite{abdollahpouri2021user} used calibration to diversify items based on popularity, tailoring the mix of mainstream and niche content to match users’ interest for each. These calibration techniques aim to align the composition of recommendation sets with the historical distribution of content types consumed by the user. 

\begin{figure}
    \centering
    \includegraphics[width=0.95\linewidth]{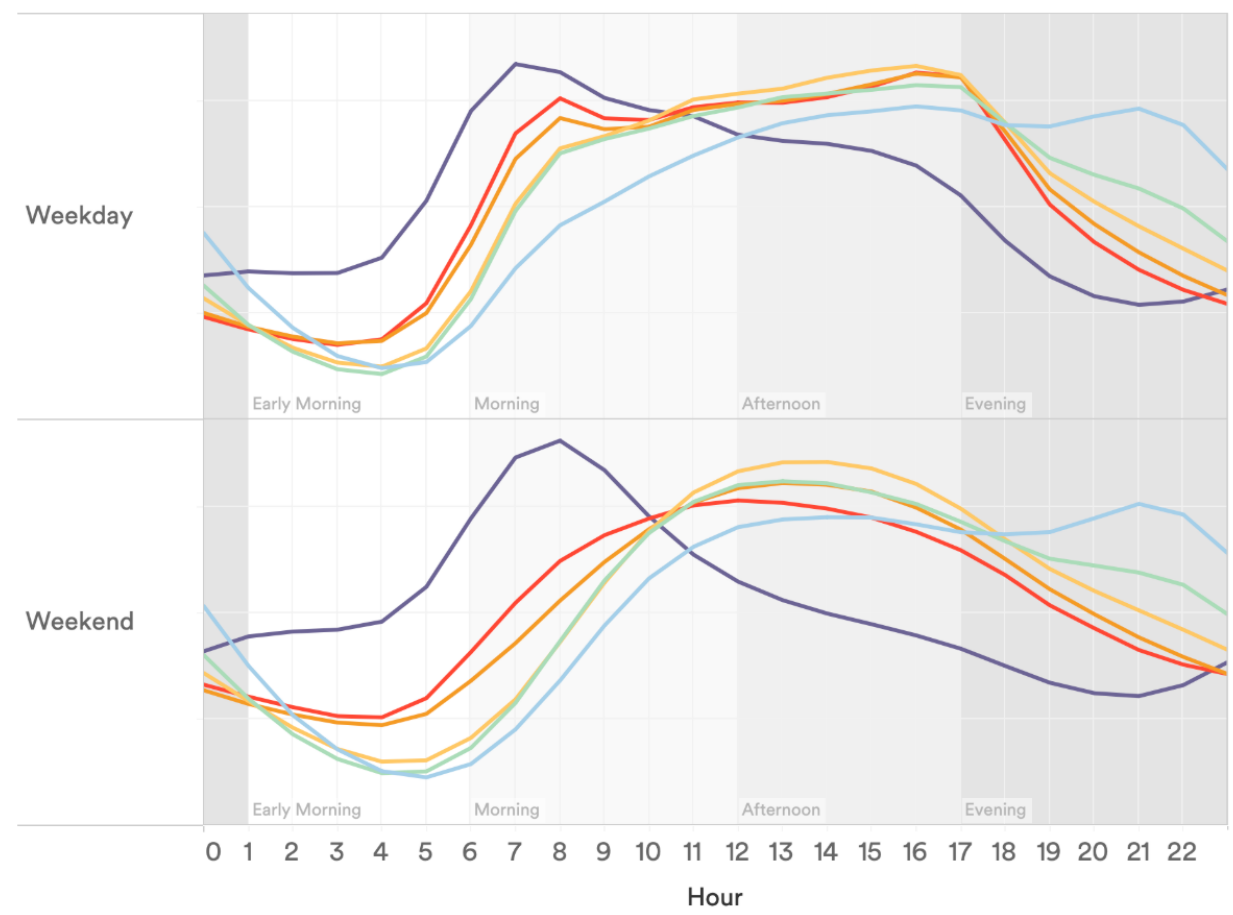}
    \caption{Illustration of dynamic content-type user preferences. Each color represents a different user cohort. The y-axis is the share of total consumption of podcasts at Spotify.}
    \label{fig:pods}
\end{figure}

While this approach can improve a user’s exposure to different content types~\cite{kaya2019comparison}, it is limited by the assumption that users' preferences are fixed. As a result, it fails to accurately capture the complex dynamics of users’ short-term and long-term interests, which can depend on a myriad of factors. For example, Figure~\ref{fig:pods} shows the daily and weekly fluctuations in podcast versus music consumption across different user cohorts at Spotify. 

In this paper, we present an approach to calibrated recommendations using a (neural) contextual bandit that learns to prescribe the optimal distribution of content types across a slate for each user, based on their current context. We apply this formulation to the calibration of music and podcasts on the Spotify Home page, where we show that it achieves superior performance in both offline evaluations and online tests compared to methods based solely on prior user preferences (e.g. estimated based on the last 90 days of interactions) or fixed target content distributions defined by business objectives~\cite{lichtenberg2024ranking}. The system described is currently deployed at Spotify and has led to increased user engagement as well as higher consumption of less mainstream content types, such as podcasts. 

Among previous work \cite{ashkan2015optimal, cai2005improving, qin2013promoting, sha2016framework}, \cite{zhao2021rabbit} also study the temporal evolution of a user's interest and calibrates recommendations based on a predicted distribution. While their approach relies on a sequential model that tracks changing interests, our solution emphasizes contextual awareness and leverages the effectiveness of contextual bandits~\cite{lattimore2020bandit} to balance exploration and exploitation~\cite{langford2007epoch, li2010contextual}.


\section{Content Type Calibration}

We now formally describe the content type calibration problem and present our proposed solution. 

\subsection{Problem Setting}
The goal of this work is to construct personalized home pages composed of different shelves for each user. Shelves can represent music, podcasts, or audiobooks. We aim to rank shelves of different content types $c$ into a slate of length $N$. 

Let $p(c|i)$ denote the content-type distribution for each shelf $i=1, \dots, N$. Our goal is to select the optimal set $I^*$ of $N$ recommended shelves using the {\it maximum marginal relevance} criterion~\cite{carbonell1998use}:
\begin{equation}
    \mathcal I^* = \argmax_{I: |I| = N} (1 - \lambda) \cdot s(I) - \lambda \cdot C_{\text{KL}}(p, I),
    \label{MMR}
\end{equation}
where $\lambda\in[0,1]$ controls the trade-off between relevance and calibration. The relevance score $s(I)$ is the sum of predicted scores $s(i)$ for each shelf $i\in I$, as given by the recommender system. Calibration is measured by the Kullback-Leibler (KL) divergence~\cite{cover1999elements}: 
\[
C_{\text{KL}}(p, q(I)) := \sum_c p(c|u)\log\frac{p(c|u)}{q(c|I)},
\]
where $p(c|u)$ is the target content-type distribution for user $u$, and $q(c|I)$ is the empirical rank-weighted distribution of content types in the recommended set:
\begin{equation}
q(c|I) := \frac{\sum_{i\in I}w_{r(i)}p(c|i)}{\sum_{i\in I} w_{r(i)}}.
\label{q}
\end{equation}
Here, $w_{r(i)}$ denotes the weight of shelf $i$ in the list, which may be uniform or depend on its rank $r(i)$,  potentially defined based on popular ranking metrics like Mean Reciprocal Rank (MRR) or normalized Discounted Cumulative Gain (nDCG).

In previous work, the target distribution $p(c|u)$ has been defined either by product-level requirements---such as global exposure goals for different content types~\cite{lichtenberg2024ranking}---or estimated from the user’s historical interactions~\cite{steck2018calibrated} as proposed by Steck:
\begin{equation}
p(c|u) = \frac{\sum_{i\in\mathcal{H}}v_{u,i}p(c|i)}{\sum_{i\in\mathcal{H}} v_{u,i}},
\label{p_cu}
\end{equation}
where $\mathcal{H}$ is the set of shelves previously interacted with by user $u$ over a certain time periods, and $v_{u,i}>0$ is a weight representing the strength of that interaction. For example, in streaming services, this weight could be based on factors such as the stream duration.

\begin{table*}[h]
    \centering
    \begin{tabular}{l||cc}
        \toprule
        \textbf{Algorithm} & \textbf{Podcast} & \textbf{Overall accuracy (music + podcast)} \\
         \midrule
         \textbf{CB (MLP) vs SC (7 days)} & ${+35\%}$ &  ${+10.2\%}$ \\
         \textbf{CB (MLP) vs SC (90 days)} & ${+25.6\%}$ &  ${+7.2\%}$  \\
         \textbf{CB (MLP) vs MB} & ${+16.6\%}$ & ${+2.8\%}$  \\
        \bottomrule
    \end{tabular}
    \caption{Offline evaluation metrics for different calibration approaches}
    \label{tab:performance}
\end{table*}

\begin{table*}[h]
\centering
 \begin{tabular}{c|| c c c c} 
 \midrule
  {\bf Model} & {\bf Podcast i2s} & {\bf Overall i2s} & {\bf Consumption} & {\bf Activity}\\ [0.5ex] 
 \midrule
{\bf CB (MLP)} & { +36.6\%} & { +3.93\% } & { +1.28\%} & { +1.54\%}\\ [1ex] 
 
\hline 
 \end{tabular}
 \caption{Online A/B experiment results vs SC (90 days)}
 \label{tab:online_performance}
\end{table*}

\subsection{Contextual Bandit Approach}

In this paper, we pose the problem of learning $p(c|u)$ as  learning with bandit feedback~\cite{swaminathan2015counterfactual, jeunen2021pessimistic, ma2019imitation}, whereby supervised learning is conducted using logs of user interactions, where 3 weeks of user data is collected and two weeks is used for training the model and the last week is used for evaluations.

In our framework, the dataset consists of i.i.d. triplets $(x_k, p_k, r_k)$, for $k=1,\dots, K$, generated through the following process:

\begin{enumerate}
    \item {\bf Context observation:} Observe a user $u_k\in\mathcal{U}$ and a set of candidate slates $\mathcal{I}_k$, along with their feature vectors $x_{k,I}$ for $I\in\mathcal{I}_k$. The vector $x_{k,I}$ gathers information of both the user and the slate (arm or action), and is referred to as the context. For simplicity, we denote this vector as $x_k$. 
    \item {\bf Action selection:} Select a content-type probability distribution $p_k$ from a discrete set $\mathcal{P}$, and construct a slate $I_k\in\mathcal{I}_k$ of size $N$ by solving Equation (\ref{MMR}) with $p=p_k$. Choosing $p_k$ is thus equivalent to choosing $I_k$. We assume this selection is made by a logging policy $\mu(p|x)$.  
    \item {\bf Reward observation:} A binary reward $r_k=h(x_k)\in\{0,1\}$ is recorded for the specific $p_k$, where $r_k=1$ if the user $u_k$ engages with (e.g. streams) any shelf $i \in I_k$, and 0 otherwise.
\end{enumerate}

The objective is to learn a policy $\pi(p|x)$ that maximizes the expected reward $\mathbb{E}_{x, p\sim\pi(p|x)}[r]$. Since the reward is binary, the optimal policy $\pi(p|x)$ is the one that selects, for the given context, a slate with a content-type distribution that maximizes the likelihood of user engagement with the slate; ties are broken at random.

\subsubsection{Learning Model}
To learn the reward function $h$, we use a fully connected neural network (MLP) with two hidden layers (sizes 256 and 64), ReLU activations, and a binary cross-entropy loss function. We also add Dropout of 0.1 to each layer for regularization. Input features include user demographics and past consumption behavior across content types aggregated over various time windows, content representations, contextual signals (e.g., device type). We learn embeddings for categorical features such as country to learn a richer representation of similar items. We encode temporal signals (e.g., time of day and day of week) by cosine and sine transformations for cyclic representations. 

\subsubsection{Exploration Policy} 
We use an $\epsilon$-greedy strategy: with a small probability of 1.5\%, we sample $p_k\in\mathcal{P}$ from the logging policy $\mu(p|x)$. We experimented with both a uniform distribution and a truncated Gaussian centered around $p(c|u)$ as defined in Equation (\ref{p_cu}). The latter yielded more positive training samples and improved model performance. Intuitively, this is because  uniform exploration often generates slates that deviate too far from user preferences and result  in  more negative outcomes, while the Gaussian approach explores around the users' interests.

\subsubsection{Optimal List Construction} Solving the combinatorial optimization problem in Equation (\ref{MMR}) is generally NP-hard. We adopt the greedy algorithm described in~\cite{shinohara2014submodular, steck2018calibrated}, which constructs $I^*$ iteratively: starting from an empty set, shelves are added one at a time to maximize the objective. Alternatively, the optimal list could be constructed more efficiently by solving a max flow optimization problem~\cite{abdollahpouri2023calibrated}. In addition, instead of the KL divergence, one could use total variation~\cite{chambolle2004algorithm}, as in the mixed-integer programming approach of~\cite{seymen2021constrained}. However, both the approaches proposed in ~\cite{abdollahpouri2023calibrated} and ~\cite{seymen2021constrained} are significantly more time consuming and may not be applicable of Spotify's Home page construction for more than 600 Million users.

\section{Evaluation}
We now present the results from both offline and online evaluations. These evaluations pertain to the Spotify Home page, where recommendation items are shelves, and the content type of each shelf include music or podcast. 

\subsection{Offline Evaluations}
Because rewards are only observed for the chosen content-type distribution, we rely on off-policy evaluation~\cite{strehl2010learning, li2011unbiased} to assess the effectiveness of different approaches. Specifically, we estimate various performance metrics for the visible portion of the Spotify Home page (i.e., without scrolling) using data collected over a 7-day period. We apply Inverse Propensity Weighting~\cite{horvitz1952generalization, bottou2013counterfactual}, with probability ratio capping to mitigate the increased variance \cite{ionides2008truncated}. 

We compare our contextual bandit approach (CB) against: 

\begin{itemize}
        \item {\bf Steck's calibration approach (SC)} from~\cite{steck2018calibrated}, where the user target distribution as defined in Equation (\ref{p_cu}), is computed based on short-term (last 7 days) or long-term (last 90 days) historical user preferences. 
        \item {\bf Blending algorithm (MB)} from~\cite{lichtenberg2024ranking}, where the recommendation list is constructed by sequentially sampling from a fixed multinomial distribution over content types.
\end{itemize}

To evaluate the quality of the generated list of shelves, we measure the accuracy of the recommended shelf that appears  on the first position on Home---specifically, just below the shortcuts links at the top---using Precision@1. For example, if our model recommends a music shelf in that position and the user streams music during the session, it is counted as a hit. The same logic applies to podcast recommendations. In addition to content-specific precision, we also report overall accuracy, which captures the model's performance irrespective of content type.


As shown in Table~\ref{tab:performance}, our approach outperformed both baselines in terms of podcast precision and overall accuracy (music performance was neutral), demonstrating its effectiveness in surfacing relevant podcast content. 



\subsection{Online A/B Testing}
We conducted an online A/B test in which the control variant was the $90$-day SC, as it outperformed the $7$-day variant in offline evaluation, and was therefore selected as a stronger baseline. As shown in Table~\ref{tab:online_performance}, our proposed methodology yielded consistent improvements across several key engagement metrics. These include impression efficiency measured by impression-to-stream (i2s) ratios, homepage activity (measured by fraction of users initiating a stream from the homepage) and overall consumption (measured by overall minutes streamed on the platform). 

Based on the positive impact across these dimensions, the system has been successfully deployed at Spotify as of March 2025.
\FloatBarrier 

\section{Conclusion and future work}

We developed a contextual bandit approach to learn content type calibration within a list of recommendations, enabling dynamic adaptation to users’ evolving and diverse preferences. This method is particularly suited to platforms like Spotify, where users engage with varied content types, such as music and podcasts, in rapidly changing contexts. Our experimental results demonstrate that this adaptive calibration method significantly outperforms baseline strategies, including those that rely on fixed content type distributions or calibrations estimated solely from historical user behavior. These baselines often fail to account for real-time changes in user interest or intent, leading to suboptimal personalization.

Our reward formulation focuses on short-term engagement. As future work, we plan to explore reward signals that incorporate multiple objectives and aim to optimize a combination of long-term user satisfaction~\cite{wu2017returning, zhang2021counterfactual, tang2023reward} and business value~\cite{jannach2019measuring}. We also intend to investigate the use of sequential signals for intent prediction, use of upper confidence bound (UCB)-based exploration~\cite{li2010contextual, zhou2020neural}, which has been shown to outperform  $\epsilon$-greedy strategies in many settings.

\bibliographystyle{ACM-Reference-Format}
\balance
\bibliography{references}


\begin{thebibliography}{34}


\ifx \showCODEN    \undefined \def \showCODEN     #1{\unskip}     \fi
\ifx \showDOI      \undefined \def \showDOI       #1{#1}\fi
\ifx \showISBNx    \undefined \def \showISBNx     #1{\unskip}     \fi
\ifx \showISBNxiii \undefined \def \showISBNxiii  #1{\unskip}     \fi
\ifx \showISSN     \undefined \def \showISSN      #1{\unskip}     \fi
\ifx \showLCCN     \undefined \def \showLCCN      #1{\unskip}     \fi
\ifx \shownote     \undefined \def \shownote      #1{#1}          \fi
\ifx \showarticletitle \undefined \def \showarticletitle #1{#1}   \fi
\ifx \showURL      \undefined \def \showURL       {\relax}        \fi
\providecommand\bibfield[2]{#2}
\providecommand\bibinfo[2]{#2}
\providecommand\natexlab[1]{#1}
\providecommand\showeprint[2][]{arXiv:#2}

\bibitem[\protect\citeauthoryear{Abdollahpouri, Mansoury, Burke, Mobasher, and Malthouse}{Abdollahpouri et~al\mbox{.}}{2021}]%
        {abdollahpouri2021user}
\bibfield{author}{\bibinfo{person}{Himan Abdollahpouri}, \bibinfo{person}{Masoud Mansoury}, \bibinfo{person}{Robin Burke}, \bibinfo{person}{Bamshad Mobasher}, {and} \bibinfo{person}{Edward Malthouse}.} \bibinfo{year}{2021}\natexlab{}.
\newblock \showarticletitle{User-centered evaluation of popularity bias in recommender systems}. In \bibinfo{booktitle}{\emph{Proceedings of the 29th ACM conference on user modeling, adaptation and personalization}}. \bibinfo{pages}{119--129}.
\newblock


\bibitem[\protect\citeauthoryear{Abdollahpouri, Nazari, Gain, Gibson, Dimakopoulou, Anderton, Carterette, Lalmas, and Jebara}{Abdollahpouri et~al\mbox{.}}{2023}]%
        {abdollahpouri2023calibrated}
\bibfield{author}{\bibinfo{person}{Himan Abdollahpouri}, \bibinfo{person}{Zahra Nazari}, \bibinfo{person}{Alex Gain}, \bibinfo{person}{Clay Gibson}, \bibinfo{person}{Maria Dimakopoulou}, \bibinfo{person}{Jesse Anderton}, \bibinfo{person}{Benjamin Carterette}, \bibinfo{person}{Mounia Lalmas}, {and} \bibinfo{person}{Tony Jebara}.} \bibinfo{year}{2023}\natexlab{}.
\newblock \showarticletitle{Calibrated recommendations as a minimum-cost flow problem}. In \bibinfo{booktitle}{\emph{Proceedings of the Sixteenth ACM International Conference on Web Search and Data Mining}}. \bibinfo{pages}{571--579}.
\newblock


\bibitem[\protect\citeauthoryear{Ashkan, Kveton, Berkovsky, and Wen}{Ashkan et~al\mbox{.}}{2015}]%
        {ashkan2015optimal}
\bibfield{author}{\bibinfo{person}{Azin Ashkan}, \bibinfo{person}{Branislav Kveton}, \bibinfo{person}{Shlomo Berkovsky}, {and} \bibinfo{person}{Zheng Wen}.} \bibinfo{year}{2015}\natexlab{}.
\newblock \showarticletitle{Optimal Greedy Diversity for Recommendation}. In \bibinfo{booktitle}{\emph{IJCAI}}, Vol.~\bibinfo{volume}{15}. \bibinfo{pages}{1742--1748}.
\newblock


\bibitem[\protect\citeauthoryear{Bottou, Peters, Qui{\~n}onero-Candela, Charles, Chickering, Portugaly, Ray, Simard, and Snelson}{Bottou et~al\mbox{.}}{2013}]%
        {bottou2013counterfactual}
\bibfield{author}{\bibinfo{person}{L{\'e}on Bottou}, \bibinfo{person}{Jonas Peters}, \bibinfo{person}{Joaquin Qui{\~n}onero-Candela}, \bibinfo{person}{Denis~X Charles}, \bibinfo{person}{D~Max Chickering}, \bibinfo{person}{Elon Portugaly}, \bibinfo{person}{Dipankar Ray}, \bibinfo{person}{Patrice Simard}, {and} \bibinfo{person}{Ed Snelson}.} \bibinfo{year}{2013}\natexlab{}.
\newblock \showarticletitle{Counterfactual reasoning and learning systems: The example of computational advertising}.
\newblock \bibinfo{journal}{\emph{The Journal of Machine Learning Research}} \bibinfo{volume}{14}, \bibinfo{number}{1} (\bibinfo{year}{2013}), \bibinfo{pages}{3207--3260}.
\newblock


\bibitem[\protect\citeauthoryear{Cai-Nicolas}{Cai-Nicolas}{2005}]%
        {cai2005improving}
\bibfield{author}{\bibinfo{person}{Ziegler Cai-Nicolas}.} \bibinfo{year}{2005}\natexlab{}.
\newblock \showarticletitle{Improving recommendation lists through topic diversification}. In \bibinfo{booktitle}{\emph{WWW'05: Proceedings of the 14th international conference on World Wide Web}}. ACM, \bibinfo{pages}{22--32}.
\newblock


\bibitem[\protect\citeauthoryear{Carbonell and Goldstein}{Carbonell and Goldstein}{1998}]%
        {carbonell1998use}
\bibfield{author}{\bibinfo{person}{Jaime Carbonell} {and} \bibinfo{person}{Jade Goldstein}.} \bibinfo{year}{1998}\natexlab{}.
\newblock \showarticletitle{The use of MMR, diversity-based reranking for reordering documents and producing summaries}. In \bibinfo{booktitle}{\emph{Proceedings of the 21st annual international ACM SIGIR conference on Research and development in information retrieval}}. \bibinfo{pages}{335--336}.
\newblock


\bibitem[\protect\citeauthoryear{Castells, Hurley, and Vargas}{Castells et~al\mbox{.}}{2021}]%
        {castells2021novelty}
\bibfield{author}{\bibinfo{person}{Pablo Castells}, \bibinfo{person}{Neil Hurley}, {and} \bibinfo{person}{Saul Vargas}.} \bibinfo{year}{2021}\natexlab{}.
\newblock \showarticletitle{Novelty and diversity in recommender systems}.
\newblock In \bibinfo{booktitle}{\emph{Recommender systems handbook}}. \bibinfo{publisher}{Springer}, \bibinfo{pages}{603--646}.
\newblock


\bibitem[\protect\citeauthoryear{Chambolle}{Chambolle}{2004}]%
        {chambolle2004algorithm}
\bibfield{author}{\bibinfo{person}{Antonin Chambolle}.} \bibinfo{year}{2004}\natexlab{}.
\newblock \showarticletitle{An algorithm for total variation minimization and applications}.
\newblock \bibinfo{journal}{\emph{Journal of Mathematical imaging and vision}}  \bibinfo{volume}{20} (\bibinfo{year}{2004}), \bibinfo{pages}{89--97}.
\newblock


\bibitem[\protect\citeauthoryear{Cover}{Cover}{1999}]%
        {cover1999elements}
\bibfield{author}{\bibinfo{person}{Thomas~M Cover}.} \bibinfo{year}{1999}\natexlab{}.
\newblock \bibinfo{booktitle}{\emph{Elements of information theory}}.
\newblock \bibinfo{publisher}{John Wiley \& Sons}.
\newblock


\bibitem[\protect\citeauthoryear{Horvitz and Thompson}{Horvitz and Thompson}{1952}]%
        {horvitz1952generalization}
\bibfield{author}{\bibinfo{person}{Daniel~G Horvitz} {and} \bibinfo{person}{Donovan~J Thompson}.} \bibinfo{year}{1952}\natexlab{}.
\newblock \showarticletitle{A generalization of sampling without replacement from a finite universe}.
\newblock \bibinfo{journal}{\emph{Journal of the American statistical Association}} \bibinfo{volume}{47}, \bibinfo{number}{260} (\bibinfo{year}{1952}), \bibinfo{pages}{663--685}.
\newblock


\bibitem[\protect\citeauthoryear{Ionides}{Ionides}{2008}]%
        {ionides2008truncated}
\bibfield{author}{\bibinfo{person}{Edward~L Ionides}.} \bibinfo{year}{2008}\natexlab{}.
\newblock \showarticletitle{Truncated importance sampling}.
\newblock \bibinfo{journal}{\emph{Journal of Computational and Graphical Statistics}} \bibinfo{volume}{17}, \bibinfo{number}{2} (\bibinfo{year}{2008}), \bibinfo{pages}{295--311}.
\newblock


\bibitem[\protect\citeauthoryear{Jannach and Jugovac}{Jannach and Jugovac}{2019}]%
        {jannach2019measuring}
\bibfield{author}{\bibinfo{person}{Dietmar Jannach} {and} \bibinfo{person}{Michael Jugovac}.} \bibinfo{year}{2019}\natexlab{}.
\newblock \showarticletitle{Measuring the business value of recommender systems}.
\newblock \bibinfo{journal}{\emph{ACM Transactions on Management Information Systems (TMIS)}} \bibinfo{volume}{10}, \bibinfo{number}{4} (\bibinfo{year}{2019}), \bibinfo{pages}{1--23}.
\newblock


\bibitem[\protect\citeauthoryear{Jeunen and Goethals}{Jeunen and Goethals}{2021}]%
        {jeunen2021pessimistic}
\bibfield{author}{\bibinfo{person}{Olivier Jeunen} {and} \bibinfo{person}{Bart Goethals}.} \bibinfo{year}{2021}\natexlab{}.
\newblock \showarticletitle{Pessimistic reward models for off-policy learning in recommendation}. In \bibinfo{booktitle}{\emph{Proceedings of the 15th ACM Conference on Recommender Systems}}. \bibinfo{pages}{63--74}.
\newblock


\bibitem[\protect\citeauthoryear{Kaya and Bridge}{Kaya and Bridge}{2019}]%
        {kaya2019comparison}
\bibfield{author}{\bibinfo{person}{Mesut Kaya} {and} \bibinfo{person}{Derek Bridge}.} \bibinfo{year}{2019}\natexlab{}.
\newblock \showarticletitle{A comparison of calibrated and intent-aware recommendations}. In \bibinfo{booktitle}{\emph{Proceedings of the 13th ACM Conference on Recommender Systems}}. \bibinfo{pages}{151--159}.
\newblock


\bibitem[\protect\citeauthoryear{Klimashevskaia, Jannach, Elahi, and Trattner}{Klimashevskaia et~al\mbox{.}}{2024}]%
        {klimashevskaia2024survey}
\bibfield{author}{\bibinfo{person}{Anastasiia Klimashevskaia}, \bibinfo{person}{Dietmar Jannach}, \bibinfo{person}{Mehdi Elahi}, {and} \bibinfo{person}{Christoph Trattner}.} \bibinfo{year}{2024}\natexlab{}.
\newblock \showarticletitle{A survey on popularity bias in recommender systems}.
\newblock \bibinfo{journal}{\emph{User Modeling and User-Adapted Interaction}} \bibinfo{volume}{34}, \bibinfo{number}{5} (\bibinfo{year}{2024}), \bibinfo{pages}{1777--1834}.
\newblock


\bibitem[\protect\citeauthoryear{Langford and Zhang}{Langford and Zhang}{2007}]%
        {langford2007epoch}
\bibfield{author}{\bibinfo{person}{John Langford} {and} \bibinfo{person}{Tong Zhang}.} \bibinfo{year}{2007}\natexlab{}.
\newblock \showarticletitle{The epoch-greedy algorithm for contextual multi-armed bandits}.
\newblock \bibinfo{journal}{\emph{Advances in neural information processing systems}} \bibinfo{volume}{20}, \bibinfo{number}{1} (\bibinfo{year}{2007}), \bibinfo{pages}{96--1}.
\newblock


\bibitem[\protect\citeauthoryear{Lattimore and Szepesv{\'a}ri}{Lattimore and Szepesv{\'a}ri}{2020}]%
        {lattimore2020bandit}
\bibfield{author}{\bibinfo{person}{Tor Lattimore} {and} \bibinfo{person}{Csaba Szepesv{\'a}ri}.} \bibinfo{year}{2020}\natexlab{}.
\newblock \bibinfo{booktitle}{\emph{Bandit algorithms}}.
\newblock \bibinfo{publisher}{Cambridge University Press}.
\newblock


\bibitem[\protect\citeauthoryear{Li, Chu, Langford, and Schapire}{Li et~al\mbox{.}}{2010}]%
        {li2010contextual}
\bibfield{author}{\bibinfo{person}{Lihong Li}, \bibinfo{person}{Wei Chu}, \bibinfo{person}{John Langford}, {and} \bibinfo{person}{Robert~E Schapire}.} \bibinfo{year}{2010}\natexlab{}.
\newblock \showarticletitle{A contextual-bandit approach to personalized news article recommendation}. In \bibinfo{booktitle}{\emph{Proceedings of the 19th international conference on World wide web}}. \bibinfo{pages}{661--670}.
\newblock


\bibitem[\protect\citeauthoryear{Li, Chu, Langford, and Wang}{Li et~al\mbox{.}}{2011}]%
        {li2011unbiased}
\bibfield{author}{\bibinfo{person}{Lihong Li}, \bibinfo{person}{Wei Chu}, \bibinfo{person}{John Langford}, {and} \bibinfo{person}{Xuanhui Wang}.} \bibinfo{year}{2011}\natexlab{}.
\newblock \showarticletitle{Unbiased offline evaluation of contextual-bandit-based news article recommendation algorithms}. In \bibinfo{booktitle}{\emph{Proceedings of the fourth ACM international conference on Web search and data mining}}. \bibinfo{pages}{297--306}.
\newblock


\bibitem[\protect\citeauthoryear{Lichtenberg, Di~Benedetto, and Ruffini}{Lichtenberg et~al\mbox{.}}{2024}]%
        {lichtenberg2024ranking}
\bibfield{author}{\bibinfo{person}{Jan~Malte Lichtenberg}, \bibinfo{person}{Giuseppe Di~Benedetto}, {and} \bibinfo{person}{Matteo Ruffini}.} \bibinfo{year}{2024}\natexlab{}.
\newblock \showarticletitle{Ranking across different content types: The robust beauty of multinomial blending}. In \bibinfo{booktitle}{\emph{Proceedings of the 18th ACM Conference on Recommender Systems}}. \bibinfo{pages}{823--825}.
\newblock


\bibitem[\protect\citeauthoryear{Ma, Wang, and Narayanaswamy}{Ma et~al\mbox{.}}{2019}]%
        {ma2019imitation}
\bibfield{author}{\bibinfo{person}{Yifei Ma}, \bibinfo{person}{Yu-Xiang Wang}, {and} \bibinfo{person}{Balakrishnan Narayanaswamy}.} \bibinfo{year}{2019}\natexlab{}.
\newblock \showarticletitle{Imitation-regularized offline learning}. In \bibinfo{booktitle}{\emph{The 22nd International Conference on Artificial Intelligence and Statistics}}. PMLR, \bibinfo{pages}{2956--2965}.
\newblock


\bibitem[\protect\citeauthoryear{McNee, Riedl, and Konstan}{McNee et~al\mbox{.}}{2006}]%
        {mcnee2006being}
\bibfield{author}{\bibinfo{person}{Sean~M McNee}, \bibinfo{person}{John Riedl}, {and} \bibinfo{person}{Joseph~A Konstan}.} \bibinfo{year}{2006}\natexlab{}.
\newblock \showarticletitle{Being accurate is not enough: how accuracy metrics have hurt recommender systems}. In \bibinfo{booktitle}{\emph{CHI'06 extended abstracts on Human factors in computing systems}}. \bibinfo{pages}{1097--1101}.
\newblock


\bibitem[\protect\citeauthoryear{Qin and Zhu}{Qin and Zhu}{2013}]%
        {qin2013promoting}
\bibfield{author}{\bibinfo{person}{Lijing Qin} {and} \bibinfo{person}{Xiaoyan Zhu}.} \bibinfo{year}{2013}\natexlab{}.
\newblock \showarticletitle{Promoting Diversity in Recommendation by Entropy Regularizer}. In \bibinfo{booktitle}{\emph{IJCAI}}, Vol.~\bibinfo{volume}{13}. Citeseer, \bibinfo{pages}{2698--2704}.
\newblock


\bibitem[\protect\citeauthoryear{Seymen, Abdollahpouri, and Malthouse}{Seymen et~al\mbox{.}}{2021}]%
        {seymen2021constrained}
\bibfield{author}{\bibinfo{person}{Sinan Seymen}, \bibinfo{person}{Himan Abdollahpouri}, {and} \bibinfo{person}{Edward~C Malthouse}.} \bibinfo{year}{2021}\natexlab{}.
\newblock \showarticletitle{A constrained optimization approach for calibrated recommendations}. In \bibinfo{booktitle}{\emph{Proceedings of the 15th ACM Conference on Recommender Systems}}. \bibinfo{pages}{607--612}.
\newblock


\bibitem[\protect\citeauthoryear{Sha, Wu, and Niu}{Sha et~al\mbox{.}}{2016}]%
        {sha2016framework}
\bibfield{author}{\bibinfo{person}{Chaofeng Sha}, \bibinfo{person}{Xiaowei Wu}, {and} \bibinfo{person}{Junyu Niu}.} \bibinfo{year}{2016}\natexlab{}.
\newblock \showarticletitle{A framework for recommending relevant and diverse items}. In \bibinfo{booktitle}{\emph{IJCAI}}, Vol.~\bibinfo{volume}{16}. \bibinfo{pages}{3868--3874}.
\newblock


\bibitem[\protect\citeauthoryear{Shinohara}{Shinohara}{2014}]%
        {shinohara2014submodular}
\bibfield{author}{\bibinfo{person}{Yusuke Shinohara}.} \bibinfo{year}{2014}\natexlab{}.
\newblock \showarticletitle{A submodular optimization approach to sentence set selection}. In \bibinfo{booktitle}{\emph{2014 IEEE International conference on acoustics, speech and signal processing (ICASSP)}}. IEEE, \bibinfo{pages}{4112--4115}.
\newblock


\bibitem[\protect\citeauthoryear{Steck}{Steck}{2018}]%
        {steck2018calibrated}
\bibfield{author}{\bibinfo{person}{Harald Steck}.} \bibinfo{year}{2018}\natexlab{}.
\newblock \showarticletitle{Calibrated recommendations}. In \bibinfo{booktitle}{\emph{Proceedings of the 12th ACM Conference on Recommender Systems}}. ACM, \bibinfo{pages}{154--162}.
\newblock


\bibitem[\protect\citeauthoryear{Strehl, Langford, Li, and Kakade}{Strehl et~al\mbox{.}}{2010}]%
        {strehl2010learning}
\bibfield{author}{\bibinfo{person}{Alex Strehl}, \bibinfo{person}{John Langford}, \bibinfo{person}{Lihong Li}, {and} \bibinfo{person}{Sham~M Kakade}.} \bibinfo{year}{2010}\natexlab{}.
\newblock \showarticletitle{Learning from logged implicit exploration data}.
\newblock \bibinfo{journal}{\emph{Advances in neural information processing systems}}  \bibinfo{volume}{23} (\bibinfo{year}{2010}).
\newblock


\bibitem[\protect\citeauthoryear{Swaminathan and Joachims}{Swaminathan and Joachims}{2015}]%
        {swaminathan2015counterfactual}
\bibfield{author}{\bibinfo{person}{Adith Swaminathan} {and} \bibinfo{person}{Thorsten Joachims}.} \bibinfo{year}{2015}\natexlab{}.
\newblock \showarticletitle{Counterfactual risk minimization: Learning from logged bandit feedback}. In \bibinfo{booktitle}{\emph{International conference on machine learning}}. PMLR, \bibinfo{pages}{814--823}.
\newblock


\bibitem[\protect\citeauthoryear{Tang, Pan, Wang, and Basilico}{Tang et~al\mbox{.}}{2023}]%
        {tang2023reward}
\bibfield{author}{\bibinfo{person}{Gary Tang}, \bibinfo{person}{Jiangwei Pan}, \bibinfo{person}{Henry Wang}, {and} \bibinfo{person}{Justin Basilico}.} \bibinfo{year}{2023}\natexlab{}.
\newblock \showarticletitle{Reward innovation for long-term member satisfaction}. In \bibinfo{booktitle}{\emph{Proceedings of the 17th ACM Conference on Recommender Systems}}. \bibinfo{pages}{396--399}.
\newblock


\bibitem[\protect\citeauthoryear{Wu, Wang, Hong, and Shi}{Wu et~al\mbox{.}}{2017}]%
        {wu2017returning}
\bibfield{author}{\bibinfo{person}{Qingyun Wu}, \bibinfo{person}{Hongning Wang}, \bibinfo{person}{Liangjie Hong}, {and} \bibinfo{person}{Yue Shi}.} \bibinfo{year}{2017}\natexlab{}.
\newblock \showarticletitle{Returning is believing: Optimizing long-term user engagement in recommender systems}. In \bibinfo{booktitle}{\emph{Proceedings of the 2017 ACM on Conference on Information and Knowledge Management}}. \bibinfo{pages}{1927--1936}.
\newblock


\bibitem[\protect\citeauthoryear{Zhang, Jia, Su, Wang, Xu, and Wen}{Zhang et~al\mbox{.}}{2021}]%
        {zhang2021counterfactual}
\bibfield{author}{\bibinfo{person}{Xiao Zhang}, \bibinfo{person}{Haonan Jia}, \bibinfo{person}{Hanjing Su}, \bibinfo{person}{Wenhan Wang}, \bibinfo{person}{Jun Xu}, {and} \bibinfo{person}{Ji-Rong Wen}.} \bibinfo{year}{2021}\natexlab{}.
\newblock \showarticletitle{Counterfactual reward modification for streaming recommendation with delayed feedback}. In \bibinfo{booktitle}{\emph{Proceedings of the 44th international ACM SIGIR conference on research and development in information retrieval}}. \bibinfo{pages}{41--50}.
\newblock


\bibitem[\protect\citeauthoryear{Zhao, Zhu, and Caverlee}{Zhao et~al\mbox{.}}{2021}]%
        {zhao2021rabbit}
\bibfield{author}{\bibinfo{person}{Xing Zhao}, \bibinfo{person}{Ziwei Zhu}, {and} \bibinfo{person}{James Caverlee}.} \bibinfo{year}{2021}\natexlab{}.
\newblock \showarticletitle{Rabbit holes and taste distortion: Distribution-aware recommendation with evolving interests}. In \bibinfo{booktitle}{\emph{Proceedings of the Web Conference 2021}}. \bibinfo{pages}{888--899}.
\newblock


\bibitem[\protect\citeauthoryear{Zhou, Li, and Gu}{Zhou et~al\mbox{.}}{2020}]%
        {zhou2020neural}
\bibfield{author}{\bibinfo{person}{Dongruo Zhou}, \bibinfo{person}{Lihong Li}, {and} \bibinfo{person}{Quanquan Gu}.} \bibinfo{year}{2020}\natexlab{}.
\newblock \showarticletitle{Neural contextual bandits with ucb-based exploration}. In \bibinfo{booktitle}{\emph{International Conference on Machine Learning}}. PMLR, \bibinfo{pages}{11492--11502}.
\newblock


\end{thebibliography}

\end{document}